\documentclass[
]{ceurart}

\usepackage{listings}
\usepackage{xcolor}       

\begin{document}

\copyrightyear{2025}
\copyrightclause{Copyright for this paper by its authors. Use permitted under Creative Commons License Attribution 4.0 International (CC BY 4.0).}
\conference{5th International Workshop on Scientific Knowledge: Representation, Discovery, and Assessment, Nov 2024, Nara, Japan}

\title{Ontologies in Motion: A BFO-Based Approach to Knowledge Graph Construction for Motor Performance Research Data in Sports Science}

\author[1,2]{Sarah Rebecca Ondraszek}[%
orcid=0009-0003-7945-6704,
email=sarah-rebecca.ondraszek@fiz-karlsruhe.de,
]
\cormark[1]

\author[1]{Jörg Waitelonis}[%
orcid=0000-0001-7192-7143,
email=joerg.waitelonis@fiz-karlsruhe.de,
]

\author[3]{Katja Keller}[%
orcid=0000-0002-8491-2247,
email=katja.keller@kit.edu,
]
\author[3]{Claudia Niessner}[%
orcid=0000-0002-8491-2247,
email=claudia.niessner@kit.edu,
]

\author[1]{Anna M. Jacyszyn}[%
orcid=0000-0002-5649-536X,
email=anna.jacyszyn@fiz-karlsruhe.de,
]
\author[1,2]{Harald Sack}[%
orcid=0000-0001-7069-9804,
email=harald.sack@fiz-karlsruhe.de,
]

\address[1]{FIZ Karlsruhe – Leibniz Institute for Information Infrastructure, Eggenstein-Leopoldshafen, Germany}
\address[2]{Institute of Applied Informatics and Formal Description Methods (AIFB) of KIT, Karlsruhe, Germany}
\address[3]{Institute of Sports and Sports Science (IfSS) of KIT, Karlsruhe, Germany}

\cortext[1]{Corresponding author.}

\begin{abstract}
An essential component for evaluating and comparing physical and cognitive capabilities between populations is the testing of various factors related to human performance. As a core part of sports science research, testing motor performance enables the analysis of the physical health of different demographic groups and makes them comparable.
The Motor Research (MO|RE) data repository, developed at the Karlsruhe Institute of Technology, is an infrastructure for publishing and archiving research data in sports science, particularly in the field of motor performance research.
In this paper, we present our vision for creating a knowledge graph from MO|RE data. With an ontology rooted in the Basic Formal Ontology, our approach centers on formally representing the interrelation of plan specifications, specific processes, and related measurements. Our goal is to transform how motor performance data are modeled and shared across studies, making it standardized and machine-understandable. The idea presented here is developed within the Leibniz Science Campus ``Digital Transformation of Research'' (DiTraRe).
\end{abstract}

\begin{keywords}
  Sports science \sep
  knowledge graphs \sep
  ontologies
\end{keywords}

\maketitle

\section{Introduction}
Research has undergone significant changes under the influence of ongoing digitalization; it has affected how scientists conduct and share research, be it in the form of digital preprints \cite{xie_is_2021} or GitHub repositories \cite{escamilla_rise_2022}. Furthermore, it revolutionized the generation and processing of digital research data, influencing its analysis and dissemination in the process \cite{peng_knowledge_2023}. 

Recent advances in artificial intelligence (AI) and the increasing prevalence of large language models (LLMs) underscore the need for improvements beyond traditional approaches to data processing and structuring. This enables machine-understandability, as current datasets are often too limited to support sophisticated automated analysis with these technologies \cite{kim_developing_2019}. In this digital transformation, ontologies and knowledge graphs (KG) can serve as foundational components to organize scientific knowledge in ways that enable machine understanding, interdisciplinary integration, and serendipitous findings \cite{waitelonis_towards_2012}.

The Leibniz Science Campus ``Digital Transformation of Research'' (DiTraRe)\footnote{DiTraRe web page, \url{https://www.ditrare.de/en}} \cite{ditrare_proposal} analyzes such digitalization processes, including state-of-the-art techniques for processing and analyzing data. It also studies the influence and effects of the digital transformation across various dimensions, using an interdisciplinary approach. In the ``Exploration and Knowledge Organization'' dimension (AI4DiTraRe), the DiTraRe research team develops AI-based methods to support real-life use cases \cite{jacyszyn_ditrare_2024}. 

One of these use cases is {\it Sensitive Data in Sports Science}, within which the Motor Research data repository (MO|RE)\footnote{MO|RE web page, \url{https://www.motor-research-data.de/}} is being developed by the Karlsruhe Institute of Technology Institute of Sports and Sports Science (KIT IfSS). MO|RE offers possibilities for storing, publishing, and archiving data about physical and cognitive capabilities generated during human motor performance testing \cite{klemm_development_2024, niessner_beyond_2025}.

Sports science data overlap with healthcare data. Existing approaches in related domains have so far not provided suitable solutions to the challenges in research data for sports science identified in DiTraRe. As such, ontologies like the Physical Activity Ontology (PACO) \cite{kim_developing_2019} are domain-specific and lack upper-level ontological grounding, which is essential for systematic interoperability. They fail to capture the complex temporal and processual aspects of measurements that are crucial for semantic integration and automated reasoning in sports science and healthcare research. As \citeauthor{kirrane_privacy_2018} (\citeyear{kirrane_privacy_2018}) already identified, a lack of consideration for privacy in ontology-based information systems is prevalent. However, access control and anonymization play a central role in this type of data \cite{dlima_security_2023}. Notably, the sports science community is increasingly establishing itself as a data-driven scientific discipline with a growing interest in interoperable, semantically structured data resources. This trend reflects both the interdisciplinary nature of the field and its ambition to contribute to evidence-based practices beyond its disciplinary boundaries \cite{keller_status_2025, eberhardt_potential_2025}. As in-silico approaches begin to enter sports science, particularly in sports epidemiology, semantic technologies such as ontologies and KGs are becoming essential tools for enabling data harmonization and interoperable, machine-readable modeling. The sports science community has shown openness to these innovations, particularly in research on the prevention of childhood diseases through physical activity \cite{niessner_silico_2025}.

The MO|RE use case functions as a proof-of-concept for the approaches aspired in DiTraRe. This concerns the modeling of complex research processes (e.g., standardized test items, measurement procedures), as well as the design rationale suitable for encoding privacy aspects. 
This paper describes the ongoing work done by the DiTraRe dimension ``Exploration and Knowledge Organisation'' within the use case {\it Sensitive Data in Sports Science}. Our contribution focuses on developing a Basic Formal Ontology (BFO)-based domain ontology and KG \cite{Arp2015BFO}. The development process addresses a range of challenges in translating domain practices into formal representations and bridging gaps in vocabulary and conceptual models, as provided by domain experts. The goal is to create a balance between granularity and complexity that supports both the extensive expressivity of MO|RE data and its reusability and cross-connections for other use cases in physical activity and human motor research. In concrete practical usage, the MO|RE ontology would enable the comparison of longitudinal motor test data, for example, for sports test results in children, by accounting for variations in test processes and conditions. In concrete practical usage, the MO|RE ontology would enable the comparison of longitudinal motor test data, for example, for sports test results in children, by accounting for variations in test processes and conditions.
\section{Related Work}
The management of research data has undergone significant evolution with the digital transformation of science. As already discovered around 20 years ago and shown in publications such as the work by \citeauthor{ludascher_managing_2006} (\citeyear{ludascher_managing_2006}), traditional approaches to the management of scientific data have proven insufficient for the complex heterogeneous datasets characteristic of modern-day research.

\subsection{Ontologies and Knowledge Graphs for Scientific Knowledge Organization}
Semantic technologies have enhanced the research data lifecycle by providing formally grounded, semantically explicit representations that capture domain semantics while facilitating cross-disciplinary linking \cite{auer_improving_2020, ahmad_toward_2024}. 

Upper-level ontologies, particularly the BFO \cite{Arp2015BFO, smith_basic_2020}, provide a foundational framework for the representation of (scientific) entities and their relationships. The BFO is highly formalized, and concepts and contributions are maintained through community-agreed standards \cite{Arp2015BFO}.

With these features, the BFO provides coherent and logical representations, as well as explicit formalizations. Ontologies that extend or build on the BFO ensure a high degree of interoperability with other knowledge systems, enabling cross-disciplinary applications \cite{smith_cornucopia_2004}.
Combined with domain-specific extensions like the Information Artifact Ontology (IAO) \cite{smith_aboutness_2015}, it lays foundational concepts for the representation of (scientific) endeavors, such as plan, process, quality, or information artifact.

In a broader infrastructural context, the National Research Data Infrastructure (NFDI) initiative in Germany brings together a diverse range of disciplinary consortia to create resources that improve the management and sharing of research data across the sciences \cite{hartl_nationale_2021}. The NFDI4Culture team developed the NFDIcore ontology as a mid-level, cross-domain model, built on BFO 2020, for representing entities and processes throughout the research lifecycle. This includes individuals, organizations, projects, datasets, services, and their relations. Together, NFDIcore enables the formal representation of scientific workflows and activities \cite{bruns_nfdicore_2024}.

Further examples of applications, such as the PMD Core Ontology (PMDco), demonstrate the wide range of uses for BFO. PMDco addresses key data management issues in materials science and engineering. As a mid-level ontology aligned with BFO 2020, PMDco models core concepts related to material classes, processing techniques, structures, properties, and performance characteristics, all of which reuse existing patterns \cite{bayerlein_pmd_2024}.

Similar to PMDco and NFDIcore, with the ontology for MO|RE data, we aim to harmonize complex measurement processes and contextual data for related test items.

\subsection{Sports Science and Motor Performance}
Formal semantic approaches to organizing and representing sports science data remain underexplored. However, ontologies in related domains provide formal representations of human motion and physiological parameters. 
The OBO Foundry ontologies pertain to anatomy and physiology, enabling the integration of biomedical data. For example, the Ontology of Biomedical Investigations (OBI) describes processes that are also relevant for motor research \cite{smith_obo_2007}. 
Moreover, PACO provides concepts for the formal representation of physical activities. Classes for the effects of exercises, equipment, and programs complement these representations. It is possible to enrich physical activities with information about their frequency, regularity, intensity,  and location \cite{kim_developing_2019}.

\section{Developing the MO|RE Ontology as a BFO-Based Module}
The IfSS developed the MO|RE data repository as a platform for publishing and archiving empirical research data in the field of sports science. In the portal, users can upload and/or download anonymized, multi-dimensional datasets with (to a certain degree) standardized motor performance testing protocols, including the metadata. Data are published open access and with a DOI. 

Derived from both long-term studies and shorter testing scenarios, the protocols capture different levels of human physical and cognitive capabilities across diverse experimental scenarios. In MO|RE data, a test item is a formalized procedure (e.g. Shuttle Run Test, Sit \& Reach, 20 meter Dash), and is also described as a comprehensive assessment protocol. This enables the enactment of a concrete, measurable task with defined execution parameters, measurement criteria, and evaluation standards, thereby describing the necessary aspects for reproducibility.

Current data organization within MO|RE data follows traditional relational database principles. The datasets are provided as spreadsheet tables for the studies, together with test protocols. These tables include anthropometric data on participants (e.g., age, height, weight, and BMI) and measurement results for test items. However, the data in the form of relational databases makes it difficult to formalize the representation of complex temporal relationships and the interrelation between predefined test item structures, their executions, and the results from real-life tests. 

\subsection{Ontology Development: Requirements Analysis}
As already mentioned, research studies in empirical sports science often involve complex relationships between predefined test items and action specifications, participants and their characteristics, as well as measurement procedures (how is the weight and height measured, and what are the contextual factors?).\footnote{\url{https://motor-research-data.de/en/Testitems_en.pdf}}
Essentially, each study available in MO|RE data represents a temporal process with multiple phases, which are condensed into a spreadsheet form as a result set. In addition to these results, MO|RE data provides metadata for each study and definitions for the test items in separate files.

Central modeling challenges include capturing the hierarchical structure of test items and their inclusion in a study. For example, when categorized accordingly in MO|RE data, a fitness evaluation can consist of multiple subcategories (strength, endurance, coordination, flexibility) for a performance profile of the participants. These categories are then divided into test items that aim to assess each of these qualities. Furthermore, these evaluations encompass a temporal dimension. This dimension may involve recording a specific time point within testing sessions in the dataset.

Nonetheless, another level of importance can be attributed to the anonymization and privacy of the included data. Sensitive data in sports science is prevalent, especially when it comes to federations and interconnection to other databases, such as KonsortSWD\footnote{\url{https://www.konsortswd.de/}}. The latter provides information on social, behavioral, educational, and economic status. Within the KG, both protected and non-protected data should be represented equally, which necessitates implementing different levels of access to maintain data privacy \cite{jacyszyn_ditrare_2024}.

\subsection{BFO Foundation}
The BFO functions as upper-level ontology for MO|RE data. Essentially, it differentiates between continuants (entities that persist through time) and occurrents (processes that unfold over time), which addresses the temporal complexity inherent in motor performance research, as represented in the datasets from the MO|RE repository. Moreover, BFO defines qualities as entities that inhere in their bearers, enabling the formal representation of measurable features of participants, such as their height, weight, or results in fitness evaluations \cite{Arp2015BFO}. 
The ontology provides conceptualizations that model processes in detail, which are also applicable to study metadata and measurement protocols. Finally, the integration of the Information Artifact Ontology (IAO) \cite{smith_aboutness_2015, IAO} and the Ontology for Biomedical Investigations (OBI) \cite{smith_obo_2007} adds further patterns for representing information content entities, measurement data, and experimental procedures.

\subsection{Exemplary Modeling Based on the Handgrip}
The Handgrip is a specific type of test item that participants undergo during evaluations to measure the maximum force that the hands can exert. As part of the measurement process, a hand dynamometer is pressed as hard as possible with one hand at a time. The resulting maximum force exerted by the hand is measured by the dynamometer and recorded as the measurement result.\footnote{\url{https://www.ifss.kit.edu/more/english/248.php}}

The vision for the first version of our BFO-based MO|RE ontology\footnote{GitHub repository of the MO|RE ontology, \url{https://github.com/ISE-FIZKarlsruhe/more-ontology}} focuses on the formal representation of studies, participants, and test items, aiming to capture the essential components of motor performance research within the repository.

\begin{figure}
    \centering
    \includegraphics[width=1\linewidth]{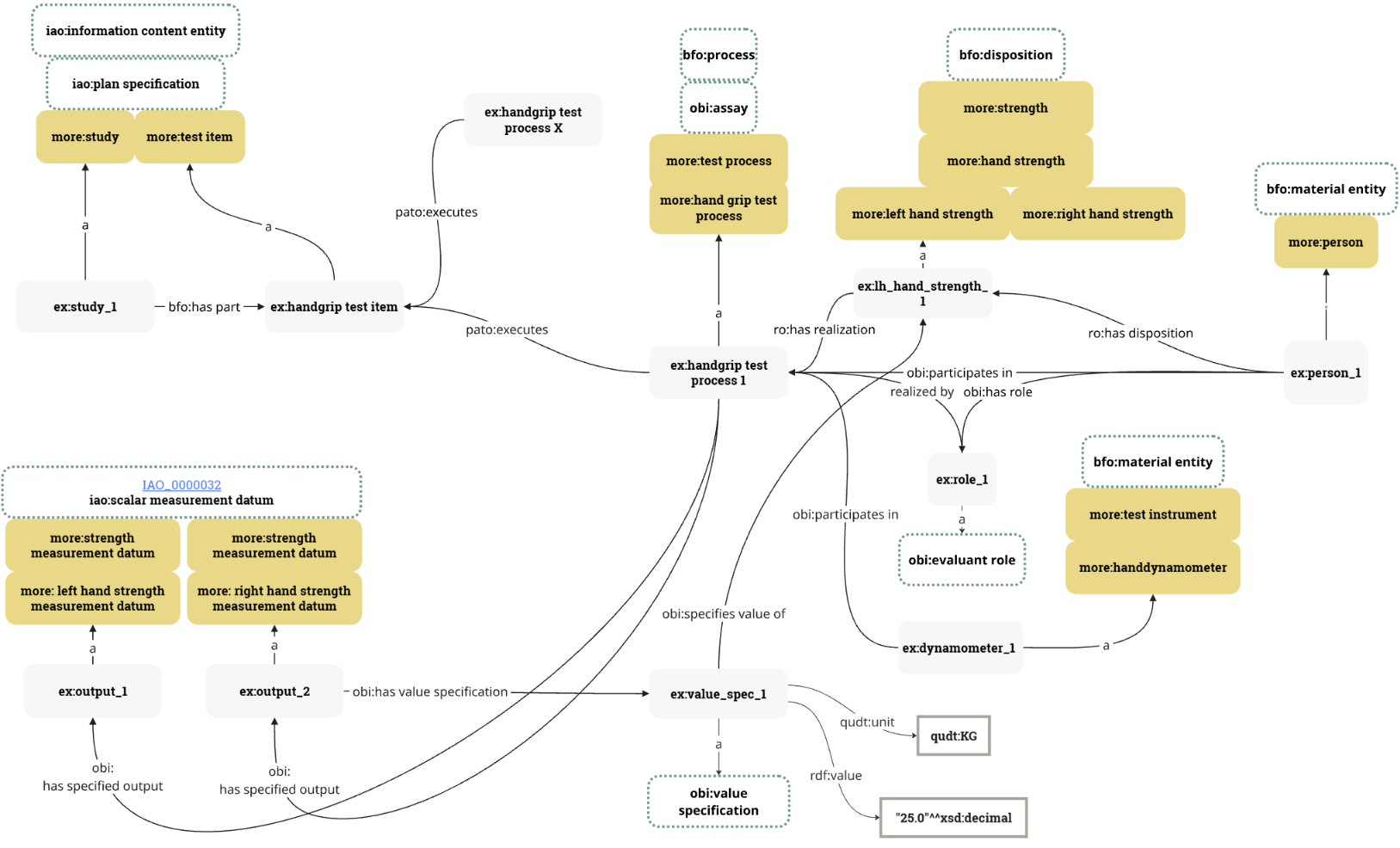}
    \caption{A part of the MO|RE ontology in development, based on an example for the Handgrip test item.}
    \label{fig:hgrip-full-vis}
\end{figure}

\subsubsection{Privacy and Data Protection for Sports Science Data}
Data featured in sports science research, such as the studies and associated test results MO|RE provides, often cover a range of personal information, be it anthropometric values or longitudinal health indicators. While the repository already anonymizes data communicated to the general public by reducing feature information, additional mechanisms are required in the ontology and KG to ensure compliance with privacy regulations (e.g., the General Data Protection Regulation given by the European Union) \cite{EstevesR24}.\footnote{\url{https://gdpr-info.eu/}}: A potential solution is to annotate properties and classes that represent identifying or health-related data (e.g., age, BMI, postal codes, etc.) with corresponding metadata for an automated distinction between sensitive and non-sensitive entities.
The implementation of required access control to ensure data privacy will be the subject of future work. Aligned with existing access restrictions and policies in systems built on semantic technologies, such as the Open Digital Rights Language (ODRL) ontology \cite{IannellaVillata2018ODRLModel}), this project considers a representation of role-based access constraints directly in the KG, so that different user groups (researchers, general public) may be granted different levels of access. This will also be reflected in the ontology design, so that sensitive values can remain under local control \cite{Decker_2024}.

\subsubsection{Studies and Test Items as Plan Specifications}
As \autoref{fig:hgrip-full-vis} shows, research studies are modeled as instances of \texttt{more:study}, a subclass of \texttt{iao:plan specification}, which is a subclass of \texttt{iao:information content entity}. This design choice reflects the understanding that a research study is a structured plan, thus an information artifact specifying a sequence of actions or processes intended to achieve certain research objectives.

The same principle applies to specifying test items in the ontology. A \texttt{more:test item} is, same as the study, an \texttt{iao:plan specification}. These test items, in essence, are action specifications (e.g., exercises like ``pushups'' or ``shuttle run'', represented as individuals) that define what participants are expected to do within the study. Each test item from a dataset is concretized in an \texttt{iao:plan}, which is accordingly realized in a concrete \texttt{more:test process}, a subclass of \texttt{obi:assay} and \texttt{bfo:process}. This is simplified with \texttt{pato:executes}. 

Accordingly, the Handgrip test item is modeled as an individual of type \texttt{more:test item}. The results are captured as the output of a measurement process, modeled as an instance of \texttt{more:handgrip test process}, which realizes the specified plan.
The actual Handgrip test process of a participant is modeled as an instance of a \texttt{more:handgrip text process}. Connected to it is the individual participant as an instance of \texttt{more:person} with the \texttt{obi:evaluant role}.

\subsubsection{Measuring Qualities and Realizations of Dispositions}

Study participants are modeled as instances of \texttt{more:person}, a subclass of \texttt{bfo:material entity} to represent their existence as continuant entities undergoing concrete assessments. All participants exhibit measurable qualities, such as anthropomorphic values measured before the evaluation (\texttt{bfo:quality}).
Following the logic of the plan specifications, participants participate in the measurement processes with corresponding roles, particularly the \texttt{obi:evaluant role}. Roles are connected to processes via \texttt{obi:has role} and realized in relations. The dispositions inherent in persons are realized through the test processes. Their values are specified via the \texttt{obi:has specified output} property, which connects a process to the measurement datum (instances of \texttt{iao:scalar measurement datum} and corresponding subclasses). This measurement datum has a value specification (\texttt{obi:value specification}), which, for the Handgrip example, is a decimal value in kilogram. This value is then used as a specification of the aforementioned disposition (\texttt{obi:specifies value of}).
As \autoref{fig:hgrip-dispositions} shows, with a shortcut based on the previously defined relations between the resources, the ontology can also express the direct connection between a test item (the Handgrip in this case) and the disposition that it measures through the process (\texttt{more:measures disposition}) via a shortcut.

\subsubsection{Scalability of Concepts in MO|RE}
The same approach can be extended to other test items from the MO|RE repository, for example, a shuttle run, in which the \texttt{more:test item} would be a \texttt{ex:shuttle run test item} and accordingly, all connected entites would be adapted (\texttt{more:shuttle run test process} with instantiations like \texttt{ex:shuttle run test process 1} for a specific execution of the test process). In the same sense, the dispositions are modeled in alignment with how a shuttle run measures the performance of a participant, e.g., measuring the time interval shown for a specific distance, or the VO2max value (maximum amount of oxygen a body can use during intense exercise).

\subsubsection{An Application Scenario}
This modeling approach enables a distinction between the planned Handgrip test, its concrete realization in a participant's test process, and the result value(s). In a concrete application scenario, in which a sports scientist is interested in finding out how motor performance in elementary pupils has changed throughout the years between 2015 and 2020, the MO|RE ontology makes it possible to compare data from multiple motor performance studies, including Handgrip strength tests conducted across different age groups and over several years, which can then be compared despite variations in protocols.
This can also be exemplified via competency questions (CQs), as applied for ontology development and evaluation in MO|RE. In this application scenario, such a CQ could be \texttt{CQ1} `How does handgrip strength vary across age groups?', or \texttt{CQ2} `Which test items were included in studies conducted between 2015–2020?' to filter out relevant triples. Both can be expressed as SPARQL queries over the MO|RE KG. 
\begin{lstlisting}[language=SPARQL, caption={\texttt{CQ1} expressed as a schematic SPARQL query for the MO|RE KG.}]
    PREFIX more: <https://w3id.org/more#>
    PREFIX obi: <http://purl.obolibrary.org/obo/OBI_>
    PREFIX iao: <http://purl.obolibrary.org/obo/IAO_>
    PREFIX xsd: <http://www.w3.org/2001/XMLSchema#>

    SELECT ?age (AVG(?strengthValue) AS ?avgStrength)
    WHERE {
      ?test a more:HandgripTestProcess ;
            obi:has_specified_output ?datum ;
            obi:has_participant ?person .
      ?datum a iao:ScalarMeasurementDatum ;
             obi:has_value_specification ?strengthValue .
      ?person more:hasAge ?age .
    }
    GROUP BY ?age
    ORDER BY ?age
    \end{lstlisting}

\section{Impact and Future Work}

The development of the MO|RE ontology and a corresponding KG is essential for the further development of motor performance test data within sports science, but especially for cooperation and collaboration with numerous external disciplines (health, education, psychology, etc.). Motor performance test data contain a wealth of information that has not yet been fully explored. The combination with other data from neighboring disciplines holds possibilities that are hardly foreseeable, for example, the relationship between motor competence and academic achievement, or the long-term health trajectories of children with low baseline motor skills. These questions require data infrastructures that go beyond single-domain perspectives. Importantly, we also emphasized the role of ontologies and knowledge graphs in promoting transparency, reproducibility, and reusability of data-core principles of open science. By structuring knowledge explicitly and making connections machine-readable, sports science stands to benefit from tools already well-established in bioinformatics and clinical research.

Additionally, based on the developments within the ontology, it is possible to define shortcuts and modules that can find application outside of this use case. This concerns shortcuts based on BFO models and, according to SWRL rules, those defined relations in particular. This plays a role in endeavors outside of DiTraRe, especially in related approaches, such as within the German National Research Data Infrastructure (NFDI), particularly in modeling complex processes and depicting data provenance.

\section{Summary}
In this paper, we introduce the vision of the MO|RE ontology, an ontology specifically designed for the needs of sports science and the MO|RE data repository. MO|RE is a platform for collecting, publishing, and sharing motor performance data. Our ontology, which is under development, is based on the BFO and reuses existing ontologies to build individual modules. The development of shared ontological frameworks in sport science is not merely a technical task; it is a strategic investment in the future of the field. It enables interdisciplinary collaboration, supports evidence-based policy, and helps build bridges between scientific discovery and real-world application, particularly in areas such as youth development, public health, and inclusive education.
The contribution presented in this paper is thus twofold: (i) to ontology engineering, it contributes a BFO-based module addressing processual modeling with privacy-aware extensions; (ii) to sports science, it provides a semantic infrastructure that standardizes motor test data.


\begin{acknowledgments}
  The Leibniz Science Campus ``Digital Transformation of Research'' (DiTraRe) is funded by the Leibniz Association (W74/2022).
\end{acknowledgments}

\section*{Declaration on Generative AI}
During the preparation of this work, the authors used DeepL and Grammarly for grammar and spelling checks. The authors reviewed and edited the content as needed and take full responsibility for the publication's content. 

\bibliography{Ondraszek_MORE_ontology_Sci-K_2025}

\appendix

\section{Shortcuts in the MO|RE Ontology}

\begin{figure}[h]
    \centering
    \includegraphics[width=\linewidth]{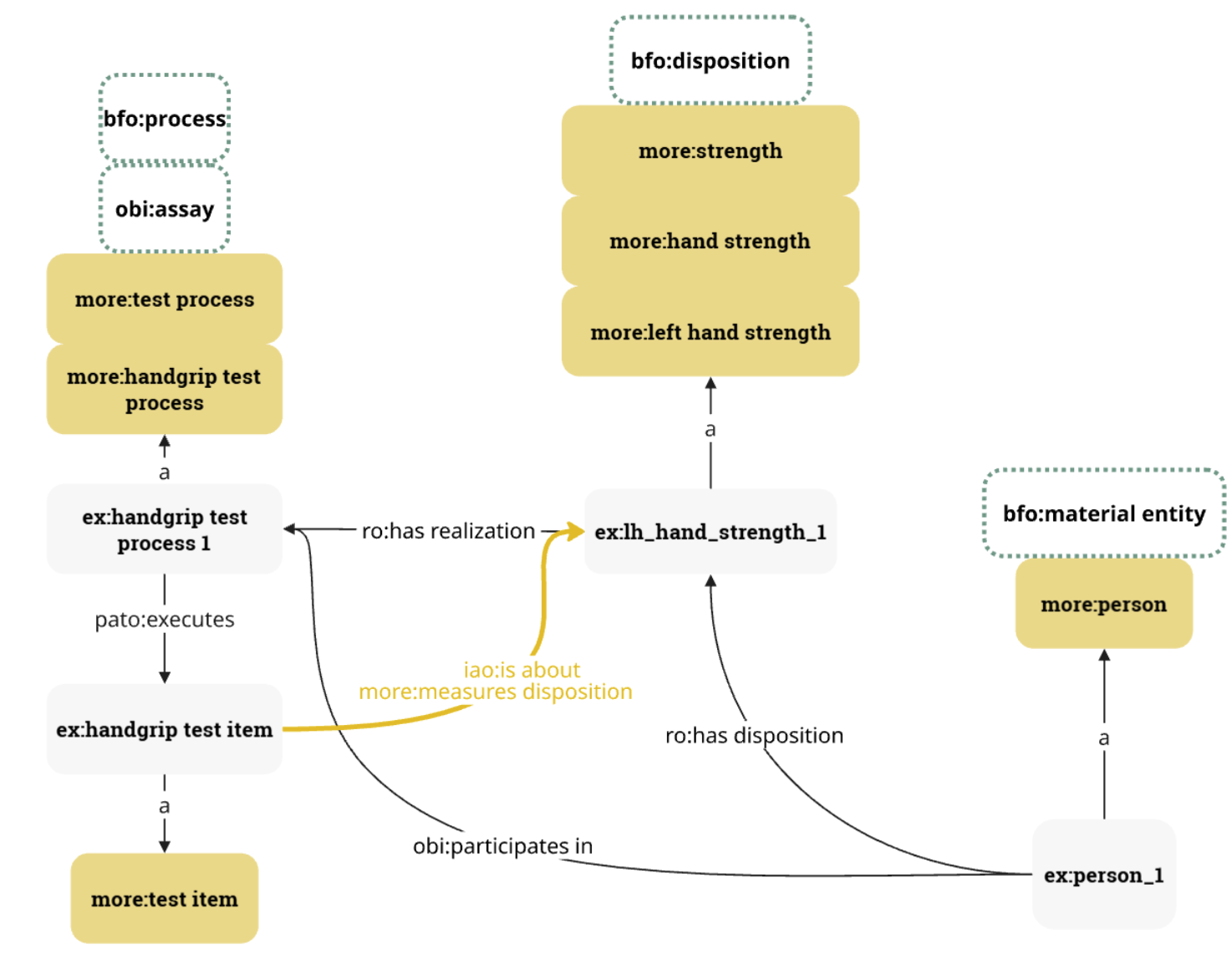}
    \caption{Dispositions module of the MO|RE ontology in development, based on an example for the Handgrip test item.}
    \label{fig:hgrip-dispositions}
\end{figure}

As \autoref{fig:hgrip-dispositions} shows, with a shortcut based on the BFO-based relations between the resources, the ontology can also express the direct connection between a test item (the Handgrip in this case) and the disposition that it measures through the process (\texttt{more:measures disposition}) via a shortcut.

\end{document}